%% file: main.tex
\documentclass[runningheads]{llncs}
\usepackage{graphicx}
\usepackage{comment}
\usepackage{amsmath,amssymb} 
\usepackage{color}
\usepackage{booktabs}
\usepackage{pifont}
\usepackage{makecell}
\usepackage{xcolor}
\usepackage{hyperref}
\hypersetup{
    colorlinks=true,
    linkcolor=magenta,
    filecolor=magenta,      
    urlcolor=magenta,
}

\usepackage[width=122mm,left=12mm,paperwidth=146mm,height=193mm,top=12mm,paperheight=217mm]{geometry}

\newcommand{\etal}{\textit{et al}.}
\newcommand{\best}[1]{\underline{\textbf{#1}}}
\newcommand{\sbest}[1]{\textbf{#1}}

\newcommand{\MI}{\text{MI}}
\newcommand{\myparagraph}[1]{\vspace{-0.0cm}\paragraph{\normalfont \normalsize \textbf{#1}}}

\makeatletter
\def\@fnsymbol#1{\ensuremath{\ifcase#1\or *\or \dagger\or \ddagger\or
   \mathsection\or \mathparagraph\or \|\or **\or \dagger\dagger
   \or \ddagger\ddagger \else\@ctrerr\fi}}
\makeatother

\definecolor{darkgreen}{rgb}{0.09, 0.45, 0.27}

\begin{document}
\pagestyle{headings}
\mainmatter
\def\ECCVSubNumber{1327}  

\title{Contrastive Learning for Weakly Supervised Phrase Grounding}

\titlerunning{Contrastive Learning for Weakly Supervised Phrase Grounding}
%
\author{Tanmay Gupta\inst{1}\thanks{Work done partly at NVIDIA}\thanks{Partly supported by ONR MURI Award N00014-16-1-2007} \and Arash Vahdat\inst{3} \and Gal Chechik\inst{2,3} \and Xiaodong Yang\inst{3} \and \\Jan Kautz\inst{3} \and Derek Hoiem\inst{1}}
\authorrunning{T. Gupta et al.}
%
\institute{University of Illinois Urbana-Champaign \and Bar Ilan University \and NVIDIA}

\maketitle

\begin{abstract}
Phrase grounding, the problem of associating image regions to caption words, is a crucial component of vision-language tasks. We show that phrase grounding can be learned by optimizing word-region attention to maximize a lower bound on mutual information between images and caption words. Given pairs of images and captions, we maximize compatibility of the attention-weighted regions and the words in the corresponding caption, compared to non-corresponding pairs of images and captions. A key idea is to construct effective negative captions for learning through language model guided word substitutions. Training with our negatives yields a $\sim10\%$ absolute gain in accuracy over randomly-sampled negatives from the training data. Our weakly supervised phrase grounding model trained on COCO-Captions shows a healthy gain of $5.7\%$ to achieve $76.7\%$ accuracy on Flickr30K Entities benchmark. Our code and project material will be available at \url{http://tanmaygupta.info/info-ground}. 


\end{abstract}

\section{Introduction}

Humans can learn from captioned images because of their ability to associate words to image regions. For instance, humans perform such word-region associations while acquiring facts from news photos, making a diagnosis from MRI scans and radiologist reports, or enjoying a movie with subtitles. 
This word-region association problem is called word or phrase \textit{grounding} and is a crucial capability needed for downstream applications like visual question answering, image captioning, and text-image retrieval.

\input{figs/teaser.tex}

Existing object detectors can detect and represent object regions in an image, and language models can provide contextualized representations for noun phrases in the caption. However, learning a mapping between these continuous, independently trained visual and textual representations is challenging in the absence of explicit region-word annotations. We focus on learning this mapping from weak supervision in the form of paired image-caption data without requiring laborious grounding annotations.


Current state-of-the-art approaches~\cite{datta2019align2ground,akbari2018semspace,rohrbach2016grounder} formulate weakly supervised phrase grounding as a multiple instance learning (MIL) problem~\cite{maron1998mil,ilse2018attentionmil}. The image can be viewed as a bag of regions. For a given phrase, all images with captions containing the phrase are treated as positive bags while remaining images are treated as negatives. Models aggregate per region features or phrase scores to construct image-level predictions that can be supervised with image-level labels in the form of phrases or captions. Common aggregation approaches include max or mean pooling, noisy-OR~\cite{fang2014cap2concepts}, and attention~\cite{datta2019align2ground,ilse2018attentionmil}. Popular training objectives include binary classification loss~\cite{fang2014cap2concepts} (whether the image contain the phrase) or caption reconstruction loss~\cite{rohrbach2016grounder} (generalization of binary classification to caption prediction) or ranking objectives~\cite{akbari2018semspace,datta2019align2ground} (do true image-caption or image-phrase pairs score higher than negative pairs). 

Fig.~\ref{fig:teaser} provides an overview of our proposed contrastive training. We propose a novel formulation of the weakly supervised phrase grounding problem as that of maximizing a lower bound on mutual information between set of region features extracted from an image and contextualized
word representations. We use pretrained region and word representations from an object detector and a language model and perform optimization over parameters of word-region attention instead of optimizing the region and word representations themselves. Intuitively, to compute mutual information with a word's representation, attention must discard nuisance regions in the word-conditional attended visual representation, thereby selecting regions that match the word. For any given word, the learned attention thus functions as a soft selection or grounding mechanism over regions. 


Since computing MI is intractable, we maximize the recently introduced InfoNCE lower bound~\cite{oord2018cpc} on mutual information. The InfoNCE bound requires a compatibility score between each caption word and the image to contrast positive image and caption word pairs with negative pairs in a minibatch. We use two objectives. 
The first objective ($\mathcal{L_\texttt{img}}$ in Fig.~\ref{fig:teaser}) contrasts a positive pair with negative pairs with the same caption word but different image regions. The second objective ($\mathcal{L_\texttt{lang}}$ in Fig.~\ref{fig:teaser}) contrasts a positive pair with negative pairs with the same image but different captions. We show empirically that sampling negative captions randomly from the training data to optimize $\mathcal{L_\texttt{lang}}$ does not yield any gains over optimizing $\mathcal{L_\texttt{img}}$ only. Instead of random sampling, we propose to use a language model to construct context-preserving negative captions by substituting a single noun word in the caption.

We design the compatibility function using a \texttt{query-key-value} attention mechanism. The \texttt{queries} and \texttt{keys}, computed from words and regions respectively, are used to compute a word-specific attention over each region which acts as a soft alignment or grounding  between words and regions. The compatibility score between regions and word is computed by comparing attended visual representation and the word representation.    

Our key contributions are: (i) a novel MI based contrastive training framework for weakly supervised phrase grounding; (ii) an InfoNCE compatibility function between a set of regions and a caption word designed for phrase grounding; 
 and (iii) a procedure for constructing context-preserving negative captions that provides $\approx10\%$ absolute gain in grounding performance.

\subsection{Related Work}
Our work is closely related to three active areas of research. We now provide an overview of prior arts in each. 

\myparagraph{Weakly Supervised Phrase Grounding.} Weakly supervised phrase localization is typically posed as a multiple instance learning (MIL) problem~\cite{maron1998mil,ilse2018attentionmil} where each image is considered as a bag of region proposals. Images whose captions mention a word or a phrase are treated as positive bags while rest of the images are treated as negatives for that word or phrase. Features or scores for a phrase or the entire caption are aggregated across all regions to make a prediction for the image. Common methods of aggregation are max or average pooling, noisy-OR~\cite{fang2014cap2concepts}, or attention~\cite{rohrbach2016grounder,ilse2018attentionmil}. With the ability to produce image-level scores for pairs of images and phrases or captions, the problem becomes an image-level fully-supervised phrase classification problem~\cite{fang2014cap2concepts} or an image-caption retrieval problem~\cite{akbari2018semspace,datta2019align2ground}. An alternatives to the MIL formulations is the approach of Ye~\etal~\cite{yeh2018concepts} which uses statistical hypothesis testing approach to link concepts detected in an image and words mentioned in the sentence. While all the above approaches assume paired image-caption data, Wang~\etal~\cite{wang2019unpairedphraseloc} recently address the problem of phrase grounding without access to image-caption pairs. Instead they assume access to a set of scene and color classifiers, and object detectors to detect concepts in the scene and use word2vec~\cite{mikolov2013w2v} similarity between concept labels and caption words to achieve grounding.

\myparagraph{MI-based Representation Learning.} Recently MI-based approaches have shown promising results on a variety representation learning problems. Computing the MI between two representations is challenging as we often have access to samples but not the underlying joint distribution that generated the samples. Thus, recent efforts rely on variational estimation of MI~\cite{alemi2016bottleneck,kim2018disentangling,belghazi2018mine,oord2018cpc}. An overview of such estimators is discussed in \cite{pool2019MIBounds,tschannen2019mutual} while the statistical limitations are reviewed in \cite{mcallester2018MIlimitations,song2019understandingMI}.

In practice, MI-based representation learning models are often trained by maximizing an estimation of MI across different \textit{transformations} of data. For example, deep InfoMax~\cite{hjelm2018deepInfoMax} maximizes MI between local and global representation using MINE~\cite{belghazi2018mine}. Contrastive predictive coding~\cite{oord2018cpc,henaff2019cpc2} inspired by noise contrastive estimation~\cite{gutmann2010nce,mnih2013nce} assumes an order in the features extracted from an image and uses summary features to predict future features. Contrastive multiview coding~\cite{tian2019cmc} maximizes MI between different color channels or data modalities while augmented multiscale Deep InfoMax~\cite{bachman2019amdim} and SimCLR~\cite{chen2020simCLR} extract views using different augmentations of data points. Since the infoNCE loss is limited by the batch size, several previous work rely on memory banks~\cite{wu2018nonparam,misra2019pretext,he2019momentum} to increase the set of negative instances.

\myparagraph{Joint Image-Text Representation Learning.} With the advances in both visual analysis and natural language understanding, there has been a recent shift towards learning representation jointly from both visual and textual domains~\cite{li2019visualbert,su2019vlbert,lu2019vilbert,sun2019videobert,tan2019lxmert,zhou2019unified,li2019unicoder,chen2019uniter,alberti2019fusion,sun2019cbt}. ViLBERT~\cite{lu2019vilbert} and LXMERT~\cite{tan2019lxmert} learn representation from both modalities using two-stream transformers, applied to image and text independently. In contrast, UNITER~\cite{chen2019uniter}, VisualBERT~\cite{li2019visualbert}, Unicoder-VL~\cite{li2019unicoder}, VL-BERT~\cite{su2019vlbert} and B2T2~\cite{alberti2019fusion} propose a unified single architecture that learns representation jointly from both domains. Our method is similar to the first group, but differs in its fundamental goal. Instead of focusing on learning a task-agnostic representation for a range of downstream tasks, we are interested in the quality of region-phrase grounding emerged by maximizing mutual information. Moreover, we rely on the language modality as a weak training signal for grounding, and we perform phrase-grounding without any further finetuning.


\section{Method}
Consider the set of region features and contextualized word representation as two multivariate random variables. Intuitively, estimating MI between them requires extracting the information content shared by these two variables. We model this MI estimation as maximizing a lower bound on MI with respect to parameters of a word-region attention model. This maximization forces the attention model to downweight regions from the image that do not match the word, and to attend to the image regions that contain the most shared information with the word representation. 



Sec.~\ref{sec:infonce_bkg} describes MI and the InfoNCE lower bound. Sec.~\ref{sec:infonce_grounding} introduces notation and InfoNCE based objective for learning phrase grounding from paired image caption data. Sec.~\ref{sec:compatibility_function} presents the design of a word-region attention based compatibility function which is part of the InfoNCE objective.   


\subsection{InfoNCE Lower Bound on Mutual Information}
\label{sec:infonce_bkg}
Let $x \in \mathcal{X}$ and $y \in \mathcal{Y}$ be random variables drawn from a joint distribution with density $p(x, y)$. The $\MI$ between $x$ and $y$ measures the amount of information that these two variables share:
\begin{equation}~\label{eq:mutual_info}
  \MI(x,y)=\mathbb{E}_{(x,y)\sim p(x,y)}\left[\log{\frac{p(x,y)}{p(x)p(y)}}\right],
\end{equation}
which is also the Kullback–Leibler Divergence from $p(x, y)$ to $p(x)p(y)$.

However, computing MI is intractable in general because it requires a complete knowledge of the joint and marginal distributions. Among the existing MI estimators, the InfoNCE~\cite{oord2018cpc} lower bound provides a low-variance estimation of MI for high dimensional data, albeit being biased~\cite{pool2019MIBounds}. The appealing variance properties of this estimator may explain its recent success in representation learning~\cite{chen2020simCLR,oord2018cpc,henaff2019cpc2,sun2019cbt}. InfoNCE defines a lower bound on MI by:  
\begin{equation}~\label{eq:infonce_bound}
  \MI(x,y) \geq \log(k) - \mathcal{L}_{k}(\theta).
\end{equation}

Here, $\mathcal{L}_k$ is the InfoNCE objective defined in terms of a compatibility function $\phi$ parametrized by $\theta$: $\phi_\theta: \mathcal{X}\times\mathcal{Y} \rightarrow \mathbb{R}$. The lower bound is computed over a mini-batch $\mathcal{B}$ of size $k$, consisting of one positive pair $(x,y) \sim p(x,y)$ and $k-1$ negative pairs $\{(x'_i,y)\}_{i=1}^{k-1}$ where ${x'} \sim p(x)$:
\begin{equation}~\label{eq:infonce_softmax}
  \mathcal{L}_{k}(\theta) = \mathbb{E}_{\mathcal{B}}\left[
    -\log\left(
    \frac{e^{\phi_{\theta}(x,y)}}{e^{\phi_{\theta}(x,y)} + \sum_{i=1}^{k-1} e^{\phi_{\theta}(x'_i,y)}}
    \right)
    \right].
\end{equation}
Oord \etal~\cite{oord2018cpc} showed that maximizing the lower bound on $\MI$ by minimizing $\mathcal{L}_k$ with respect to $\theta$ leads to a compatibility function $\phi_{\theta^*}$ that obeys 
\begin{equation}
  e^{\phi_{\theta^*}(x,y)} \propto \frac{p(x|y)}{p(x)} = \frac{p(x,y)}{p(x)p(y)},
\end{equation}
where $\theta^*$ is the optimal $\theta$ obtained by minimizing $\mathcal{L}_k$.

\subsection{InfoNCE for Phrase Grounding}~\label{sec:infonce_grounding}
Recent work~\cite{datta2019align2ground} has shown that pre-trained object detectors such as FasterRCNN~\cite{ren2015fasterrcnn} and language models such as BERT~\cite{devlin2018bert} provide rich representations in the visual and textual domains for the phrase grounding problem. Inspired by this, we aim to maximize mutual information between region features generated by an object detector and contextualized word representation extracted by a language model.

Let us denote image region features for an image by $\mathbf{R}=\{r_i\}_{i=1}^{m}$ where $m$ is the number of regions in the image with each $r_i \in \mathbb{R}^{d_r}$. Similarly, caption word representations are denoted as $\mathbf{W}=\{w_j\}_{j=1}^{n}$ where $n$ is the number of words in the caption with each word represented as $w_j \in \mathbb{R}^{d_w}$.

We maximize the InfoNCE lower bound on MI between image regions and each individual word representation denoted by $\MI(\mathbf{R},w_j)$. Thus using Eq.~\ref{eq:infonce_bound} we maximize the following lower bound: 

\begin{equation}~\label{eq:infonce_grounding}
  \sum_{j=1}^n \MI(\mathbf{R},w_j) \geq n \log(k) - \sum_{j=1}^n \mathcal{L}_{kj}(\theta).
\end{equation}
We empirically show that maximizing the lower bound in Eq.~\ref{eq:infonce_grounding} with an appropriate choice of compatibility function $\phi_{\theta}$ results in learning phrase grounding without strong grounding supervision. The following section details the design of the compatibility function. 

\input{figs/compatibility.tex}

\subsection{Compatibility Function with Attention}~\label{sec:compatibility_function}
The InfoNCE loss in our phrase grounding formulation requires a compatibility function between the \textit{set} of region feature vectors $\mathbf{R}$ and the contextualized word representation $w_j$. To define the compatibility function, we propose to use a \texttt{query-key-value} attention mechanism~\cite{vaswani2017transformer}. Specifically, we define neural modules $k_r,v_r:\mathbb{R}^{d_r}\rightarrow\mathbb{R}^{d}$ to map each image region to \texttt{keys} and \texttt{values} and $q_w, v_w:\mathbb{R}^{d_w}\rightarrow\mathbb{R}^{d}$ to compute \texttt{query} and \texttt{values} for the words. The \texttt{query} vectors for each word are used to compute the attention score for every region given a word using 
\begin{equation}
  a(r_i,w_j) = \frac{e^{s(r_i,w_j)}}
  {\sum_{i'=1}^m e^{s(r_{i'},w_j)}}, 
\end{equation}
where $s(r_i, w_j) =  q_w(w_j)^T k_r(r_i) / \sqrt{d}$. The attention scores are used as a soft selection mechanism to compute a word-specific visual representation using a linear combination of region \texttt{values}
\begin{equation}
  v_{att}(\mathbf{R},w_j)=\sum_{i=1}^m a(r_i,w_j)v_r(r_i).
\end{equation}

Finally, the compatibility function is defined as $\phi_\theta(\mathbf{R},w_j) = v_w^T(w_j)v_{att}(\mathbf{R},w_j)$, where $\theta$ refers to the parameters of neural modules $k_r,v_r,q_w, \text{and} \;v_w$, implemented using simple feed-forward MLPs. Following Eqs.~\ref{eq:infonce_softmax} \&~\ref{eq:infonce_grounding}, the InfoNCE loss for phrase grounding is defined as 
\begin{equation}~\label{eq:infonce_grounding_loss}
  \mathcal{L}_{\texttt{img}}(\theta) = \mathbb{E}_{\mathcal{B}}\left[
    -\sum_{j=1}^{n}\log\left(
    \frac{e^{\phi_{\theta}(\mathbf{R},w_j)}}{e^{\phi_{\theta}(\mathbf{R},w_j)} + \sum_{i=1}^{k-1} e^{\phi_{\theta}(\mathbf{R}'_i,w_j)}}
    \right)
    \right].
\end{equation}
which is marked using subscript \texttt{img} as negative pairs are created by replacing image regions from a positive pair with regions extracted from negative instance in the mini-batch.

\myparagraph{Remark:} We enforce compatibility between each word and all image regions using $\MI(\mathbf{R},w_j)$ in Eq.~\ref{eq:infonce_grounding}, but not between a region and all caption words ($\MI(r_i, \mathbf{W})$). This is because the words only describe part of the image, so there will be regions with no corresponding word in the caption.

\subsection{Context-Preserving Negative Captions}~\label{sec:neg_caps}
\input{figs/negatives}
The objective in Eq.~\ref{eq:infonce_grounding_loss} trains the compatibility function by contrasting positive regions-word pairs against pairs with replaced image regions. We now propose a complementary objective function that contrasts the positive pairs against negative pairs whose captions are replaced with plausible negative captions. However, extracting negative captions that are related to a captions is challenging as it requires semantic understanding of words in a caption. Here, we leverage BERT as a pretrained bidirectional language model to extract such negative captions. 


For a caption with a noun word $s$ and context $c$, we define a context-preserving negative caption as one which has the same context $c$ but a different noun $s'$ with the following properties: (i) $s'$ should be plausible in the context; and (ii) the new caption defined by the pair $(s',c)$ should be untrue for the image. For example, consider the caption \texttt{"A man is walking on a beach"} where $s$ is chosen as \texttt{"man"} and $c$ is defined by \texttt{"A [MASK] is walking on a beach"} where \texttt{[MASK]} is the token that denotes a missing word. A potential candidate for a context-preserving negative caption might be \texttt{"A woman is walking on a beach"} where $s'$ is \texttt{woman}. However, \texttt{"A car is walking on a beach"} and \texttt{"A person is walking on a beach"} are not negative captions because \texttt{car} is not plausible given the context, and the statement with \texttt{person} is still true given that the original caption is true for the image.

\myparagraph{Constructing context-preserving negative captions.} We propose to use a pre-trained BERT language model to construct context-preserving negative captions for a given true caption. Our approach for extracting such words consists of two steps: First, we feed the context $c$ into the language model to extract $30$ most likely candidates $\{s'_l\}_{l=1}^{30}$ for the masked word using probabilities $p(s'|c)$ predicted by BERT. Intuitively, these words correspond to those that fill in the masked word in caption according to BERT. However, the original masked word or its synonyms may be present in the set as well.
Thus, in the second step, we pass the original caption into BERT to compute $q(s_l'|s,c)$ which we use as a proxy for how true $(s_l',c)$ is given that $(s, c)$ is true. We re-rank the candidates using the score $\frac{p(s'|c)}{q(s'|s,c)}$ and we keep the top $25$ captions $\{(s'_l,c)\}_{l=1}^{25}$ as negatives for the original caption $(s,c)$. 

We empirically find that the proposed approach is effective in extracting context-preserving negative captions. Fig.~\ref{fig:negatives} shows a context-preserving negatives for a set of captions along with candidates that were rejected after re-ranking. Note that the selected candidates match the context and the rejected candidates are often synonyms or hypernyms of the true noun. 

\myparagraph{Training with context-preserving negative captions.} Given the context-preserving negative captions, we can train our compatibility function by contrasting the positive pairs against negative pairs with plausible negative captions. We use a loss function similar to InfoNCE to encourage higher compatibility score of an image with the true caption than any negative caption. Let $w$ and $\{w_l'\}_{l=1}^{25}$ denote the contextualized representation of the positive word $s$ and the corresponding negative noun words $\{s_l'\}_{l=1}^{25}$. The language loss is defined as
\begin{equation}~\label{eq:neg_loss}
  \mathcal{L}_{\texttt{lang}}(\theta) = \mathbb{E}_{\mathcal{B}}\left[
    -\log\left(
    \frac{e^{\phi_{\theta}(\mathbf{R}, w)}} {e^{\phi_{\theta}(\mathbf{R},w)} + \sum_{l=1}^{25} e^{\phi_{\theta}(\mathbf{R},w_{l}')}}
    \right)\right].
\end{equation}
 For captions with multiple noun words, we randomly select $s$ from the noun words for simplicity.

\subsection{Implementation Details}
\myparagraph{Regions and Visual Features.} We use the Faster-RCNN object detector provided by Anderson~\etal~\cite{anderson2017bottomuptopdown} and used for extracting visual features in the current state-of-the-art phrase grounding approach Align2Ground~\cite{datta2019align2ground}. The detector is trained jointly on Visual Genome object and attribute annotations and we use a maximum of 30 top scoring bounding boxes per image with $2048$ dimensional ROI-pooled region features.

\myparagraph{Contextualized Word Representations.} We use a pretrained BERT language model to extract $768$ dimensional contextualized word representations for each caption word. Note that BERT is trained on a text corpora using masked language model training where words are randomly replaced by a \texttt{[MASK]} token in the input and the likelihood of the masked word is maximized in the distribution over vocabulary words predicted at the output. Thus, BERT is trained to model distribution over words given context and hence suitable for modeling $p(s|c)$ defined in Sec.~\ref{sec:neg_caps} for constructing context-preserving negative captions.

\myparagraph{Query-Key-Value Networks.} We use an MLP with 1 hidden layer for each of $k_r,v_r,q_w,v_w$ for all experiments except the ablation in Fig.~\ref{fig:infonce_acc_plot}. We use BatchNorm~\cite{Ioffe2015batchnorm} and ReLU activations after the first linear layer. The hidden layer has the same number of neurons as the input dimensions of these networks which are $2048$ for $(k_r, v_r)$, and $768$ for $(q_w, v_w)$. The output layer is $384$ ($=768/2$) for all networks.

\myparagraph{Losses.} Since we only  care about grounding noun phrases, we compute $\mathcal{L}_\text{img}$ only for noun and adjective words in the captions as identified by a POS tagger instead of all caption words for computation efficiency.

\myparagraph{Optimization.} We optimize $\mathcal{L}_{img} + \mathcal{L}_{lang}$  computed over batches of 50 image-caption pairs using the ADAM optimizer~\cite{kingma2014adam} with a learning rate of $10^{-5}$. We compute $\mathcal{L}_\texttt{img}$ for each image using other images in the batch as negatives.

\myparagraph{Attention to phrase grounding.} We use the BERT tokenizer to convert captions into individual word or sub-word tokens. Attention is computed per token. For evaluation, the phrase-level attention score for each region is computed  as the maximum attention score assigned to the region by any of the tokens in the phrase. The regions are then ranked according to this phrase level score. 

\section{Experiments}
Our experiments compare our approach to state-of-the-art on weakly supervised phrase localization (Sec.~\ref{sec:sota_comparison}), ablate gains due to pretrained language representations and context-preserving negative sampling using a language model (Sec.~\ref{sec:lang_modeling}), and analyse the relation between phrase grounding performance and the InfoNCE bound that we optimize as a proxy for phrase grounding (Sec.~\ref{sec:infonce_vs_grounding}). 


\subsection{Datasets and Metrics}

We train our models on image-caption pairs from COCO training set which consists of $\sim83$K training images. We use the validation set with $\sim41$K images for part of our analysis. Each image is accompanied with 5 captions. 
For evaluation, we use the Flickr30K Entities validation set for model selection (early stopping) and test set for reporting final performance. Both sets consist of $1$K images with 5 captions each. We report two metrics:

\myparagraph{Recall@k} 
which is the fraction of phrases for which the ground truth bounding box has an IOU $\geq 0.5$ with any of the top-k predicted boxes.

\myparagraph{Pointing accuracy}
which requires the model to predict a single point location per phrase and the prediction is counted as correct if it falls within the ground truth bounding box for the phrase. Unlike recall@k, pointing accuracy does not require identifying the extent of the object. Since our model selects one of the detected regions in the image, we use use center of the selected bounding box as the prediction for each phrase for computing pointing accuracy.

\subsection{Performance on Flickr30K Entities}~\label{sec:sota_comparison}
\input{tabs/flickr.tex}
Tab.~\ref{tab:flickr} compares performance of our method to existing weakly supervised phrase grounding approaches on the Flickr30K Entities test set. A few existing approaches train on Flickr30K Entities train set and report recall@1 while recent methods use COCO train set and report pointing accuracy. Further, all approaches use different visual features making direct comparison difficult. For a fair comparison to state-of-the-art, we use Faster-RCNN trained on Visual Genome object and attribute annotations used in Align2Ground~\cite{datta2019align2ground} and report performance for models trained on either datasets on both recall and pointing accuracy metrics.

Using the same training data and visual feature architecture, our model shows a $5.7\%$ absolute gain in pointing accuracy over Align2Ground. Learning using our contrastive formulation is also quite sample efficient as can be seen by only a $2$ to $3$ points drop in performance when the model is trained on the much smaller Flickr30K Entities train set which has approximately one-third as many image-caption pairs as COCO.

\subsection{Benefits of Language Modeling}~\label{sec:lang_modeling}
\input{tabs/lang_model}
Our approach benefits from language modeling in two ways: (i) using the pretrained language model to extract contextualized word representations, and (ii) using the language model to sample context-preserving negative captions. Tab.~\ref{tab:lang_model} evaluates along both of these dimensions.

\myparagraph{Gains from pretrained word representations.} In Tab.~\ref{tab:lang_model}, {\texttt{BERT (Random)}} refers to the BERT architecture initialized with random weights and finetuned on COCO image-caption data along with parameters of the attention mechanism. \texttt{BERT (Pretrained)} refers to the off-the-shelf pretrained BERT model which is used as a contextualized word feature extractor during contrastive learning without finetuning. We observe a $\sim$10\% absolute gain in both recall@1 and pointing accuracy by using pretrained word representations from BERT. 

\myparagraph{Gains from context-preserving negative caption sampling.} Our context-preserving negative sampling has two steps. The first step is drawing negative noun candidates given the context provided by the true caption. The second step is re-ranking the candidates to filter out likely synonyms or hypernyms that are also true for the image.

First, note that randomly sampling negative captions from training data for computing $\mathcal{L}_{\text{lang}}$ performs similarly to only training using $\mathcal{L}_{\text{img}}$. Model trained with contextually plausible negatives significantly outperforms random sampling by $\geq$8\% gain in recall@1 and pointing accuracy. Excluding near-synonyms and hypernyms yields another $\sim$3 points gain in recall@1 and accuracy.

\subsection{Is InfoNCE a good proxy for learning phrase grounding?}~\label{sec:infonce_vs_grounding}
\input{figs/infonce_acc_plot.tex}

The fact that optimizing our InfoNCE objective results in learning phrase grounding is intuitive but not trivial. Fig.~\ref{fig:infonce_acc_plot} shows that maximizing the InfoNCE lower bound correlates well with phrase grounding performance on a heldout dataset. We make several interesting observations: \textbf{(i)} As training progresses (from left to right), InfoNCE lower bound (Eq.~\ref{eq:infonce_grounding}) mostly keeps increasing on the validation set. This indicates that there is no overfitting in terms of the InfoNCE bound. \textbf{(ii)}  With the increase in InfoNCE lower bound, phrase grounding performance first increases until peak performance and then decreases. This shows that the InfoNCE bound is correlated with the grounding performance but maximizing it fully does not necessarily yield the best grounding. A similar observation has been made in \cite{tian2019cmc} for representation learning. \textbf{(iii)}  The peak performance and the number of iterations needed for the best performance depends on the choice of \texttt{key-value-query} modules. One and two layer MLPs hit the peak faster and perform better than linear functions. We refer the reader to Sec.~\ref{appendix:limitations} in the appendix for a discussion of limitations of our approach.

\subsection{Qualitative Results}

Fig.~\ref{fig:qual_results} visualizes the word-region attention learned by our model. The qualitative results demonstrate the following abilities: \textbf{(i)} localizing different objects mentioned in the same caption with varying degrees of semantic relatedness, e.g., \texttt{man} and \texttt{canine} in row 1 vs. \texttt{man} and \texttt{woman} in row 3; \textbf{(ii)} disambiguation between two instances of the same object category using caption context. For example, \texttt{boy} and \texttt{another} in row 4 and bride and groom from other men and women in row 3; \textbf{(iii)} localizing object parts such as toddler's \texttt{shirt} in row 2 and instrument's \texttt{mouthpiece} in row 5; \textbf{(iv)} handling occlusion, e.g., \texttt{table} covered with toys in row 6; \textbf{(v)} handling uncommon words or categories like \texttt{ponytail} and \texttt{mouthpiece} in row 5 and \texttt{hose} in row 7. 





\section{Conclusion}
 In this work, we offer a novel perspective on weakly supervised phrase grounding from paired image-caption data which has traditionally been cast as a multiple instance learning problem. We formulate the problem as that of estimating mutual information between image regions and caption words. We demonstrate that maximizing a lower bound on mutual information with respect to parameters of a region-word attention mechanism results in learning to ground words in images. We also show that language models can be used to generate context-preserving negative captions which greatly improve learning in comparison to randomly sampling negatives from training data. 

 \input{figs/qual_results.tex}

\clearpage
%
%
\bibliographystyle{splncs04}
\bibliography{main}

\appendix

\input{appendix}
\end{document}

%% file: figs/teaser.tex
\begin{figure}[t]
    \centering
    \includegraphics[width=1\linewidth, trim={0cm, 0, 0, 0cm}, clip=True]{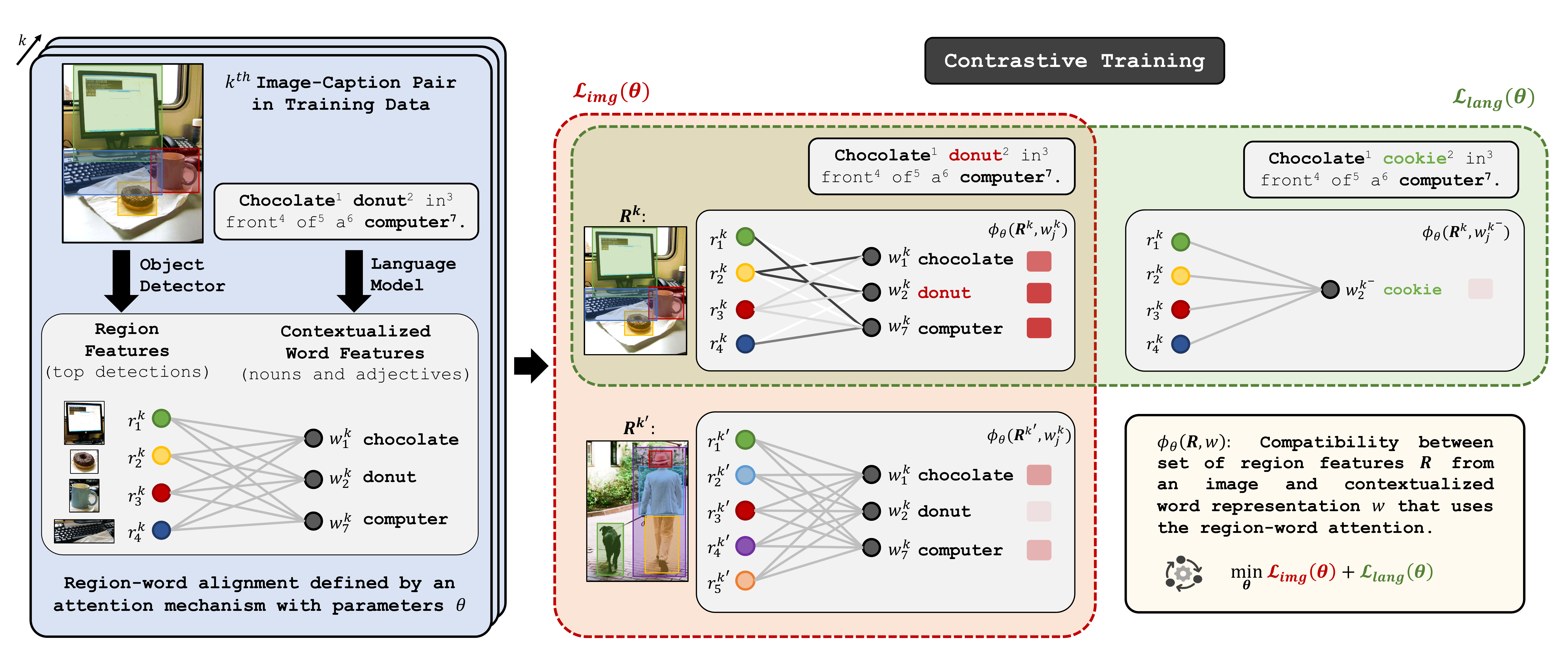}
    \caption{\textbf{Overview of our contrastive learning framework.} We begin by extracting region and word features using an object detector and a language model respectively. Contrastive learning trains a word-region attention mechanism as part of a compatibility function $\phi_{\theta}$ between the set of region features from an image and individual contextualized word representations. The compatibility function is trained to maximize a lower bound on mutual information with two losses. For a given caption word, $\color{red}{\mathcal{L}_{\texttt{img}}}$ learns to produce a higher compatibility for the true image than a negative image in the mini-batch. ${\color{darkgreen}\mathcal{L}_{\texttt{lang}}}$ learns to produce a higher compatibility of an image with a true caption-word than with a word in a negative caption. We construct negative captions by substituting a noun word like ``donut" in the true caption with contextually plausible but untrue words like ``cookie" using a language model.}
    \label{fig:teaser}
\end{figure}

%% file: figs/compatibility.tex
\begin{figure}[t]
    \centering
    \includegraphics[width=0.9\linewidth, trim={0cm, 0, 0, 1cm}, clip=True]{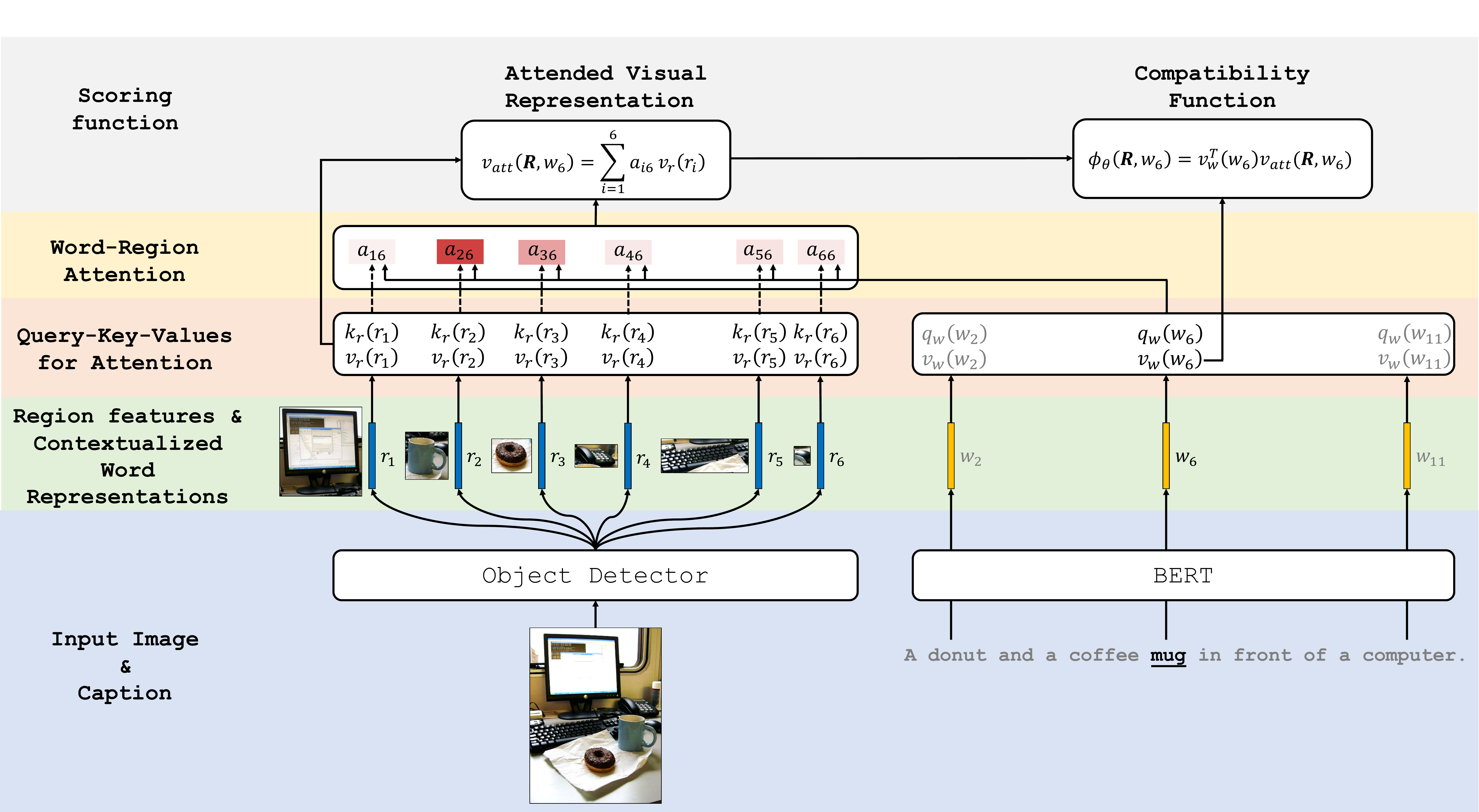}
    \caption{\textbf{Compatibility function $\phi_\theta$ with word-region attention.} The figure shows compatibility computation between the set of image regions and the word ``mug" in the caption. The compatibility function consists of learnable \texttt{query-key-value} functions $k_r,v_r,q_w,v_w$. The \texttt{query} constructed from contextualized representation of the word ``mug" is compared to \texttt{keys} created from region features to compute attention scores. The attention scores are used as weights to linearly combine \texttt{values} created from region features to construct an attended visual representation for ``mug". The compatibility is defined by the dot product of the attended visual representation and \texttt{value} representation for ``mug".}
    \label{fig:compatibility}
\end{figure}

%% file: figs/negatives.tex
\begin{figure}[t]
    \centering
    \includegraphics[width=\linewidth]{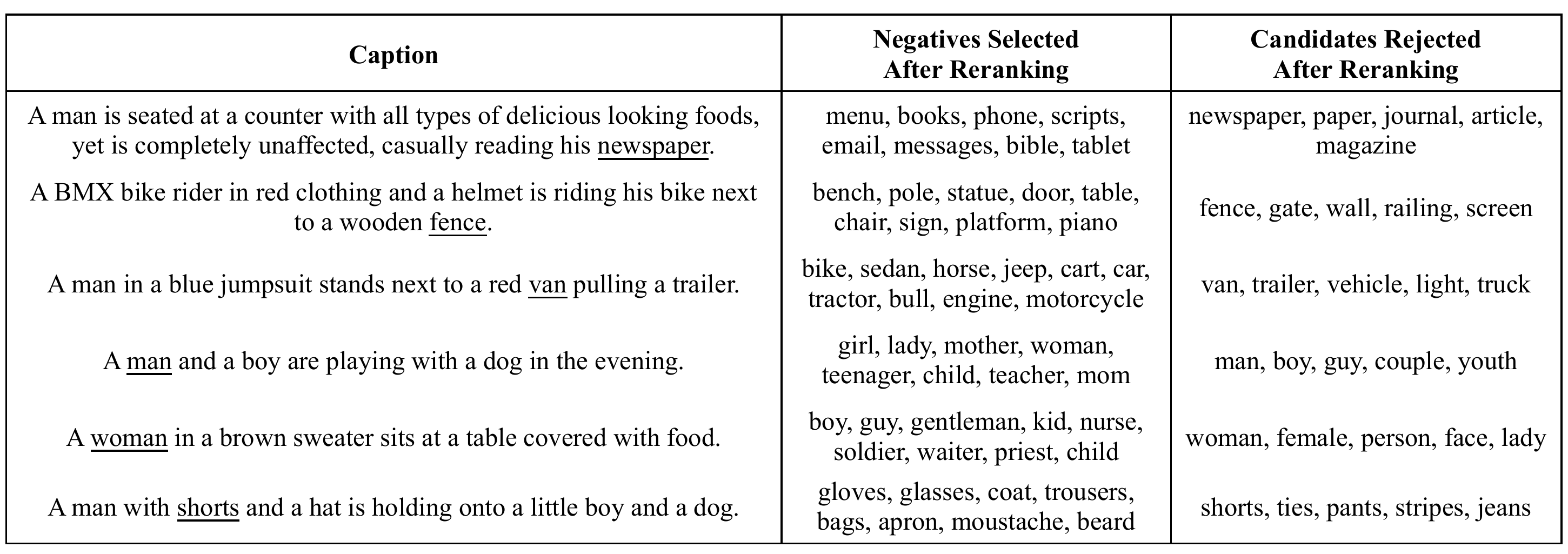}
    \caption{\textbf{Context-preserving negative captions.} We construct negative captions which share the same context as the true caption but substitute a noun word. We choose the substitute using a language model such that it is plausible in the context but we reject potential synonyms or hypernyms of the original word by a re-ranking procedure.}
    \label{fig:negatives}
\end{figure}

%% file: tabs/flickr.tex
\setlength{\tabcolsep}{2pt}
\begin{table}[t]
\scriptsize
\begin{center}
\caption{\textbf{Grounding performance on Flickr30K Entities test set}. We make our approach directly comparable to the current state-of-the-art, Align2Ground~\cite{datta2019align2ground}. The performance of older methods are reported for completeness but the use of different visual features makes direct comparison difficult.}%
\vskip -8pt%
\label{tab:flickr}%
\resizebox{\columnwidth}{!}{%
\begin{tabular}{lcccccc}%
\toprule\noalign{\smallskip}
\bf Method & \bf Training Data & \bf Visual Features & \bf R@1 & \bf R@5 & \bf R@10 & \bf Accuracy \\
\noalign{\smallskip}
\midrule
\noalign{\smallskip} 
GroundeR (2015)~\cite{rohrbach2016grounder} & Flickr30K Entities & VGG-det (VOC) & 28.94 & - & - & - \\
Yeh~\etal (2018)~\cite{yeh2018concepts} & Flickr30K Entities & VGG-cls (IN) & 22.31 & - & - & - \\
Yeh~\etal (2018)~\cite{yeh2018concepts} & Flickr30K Entities & VGG-det (VOC) & 35.90 & - & - & - \\
Yeh~\etal (2018)~\cite{yeh2018concepts} & Flickr30K Entities & YOLO (COCO) & 36.93 & - & - & - \\
KAC Net+Soft KBP (2018)~\cite{chen2018kac} & Flickr30K Entities & VGG-det (VOC) & 38.71 & - & - & - \\
\midrule
Fang~\etal (2015)~\cite{fang2014cap2concepts}    & COCO   & VGG-cls (IN) & - & -  &  - &   29.00\\
Akbari~\etal (2019)~\cite{akbari2018semspace}   & COCO    & VGG-cls (IN) & - & -  &  - &   61.66\\
Akbari~\etal (2019)~\cite{akbari2018semspace}   & COCO    & PNAS Net (IN) & - & -  &  - &   69.19\\
Align2Ground (2019)~\cite{datta2019align2ground} & COCO    & Faster-RCNN (VG) & -  & -  &  - &   71.00\\ 
\midrule
Ours & Flickr30K Entities & Faster-RCNN (VG) & \sbest{47.88} & \sbest{76.63} & \sbest{82.91} & \sbest{74.94} \\
Ours    & COCO & Faster-RCNN (VG) & \best{51.67} & \best{77.69} & \best{83.25} & \best{76.74} \\
\bottomrule
\end{tabular}}
\end{center}
\end{table}
\setlength{\tabcolsep}{1.4pt}

%% file: tabs/lang_model.tex
\setlength{\tabcolsep}{2pt}
\begin{table}[t]
\scriptsize
\begin{center}
\caption{\textbf{Benefits of language modeling.} The first two rows show the gains due to pretrained language representations. The next three rows show gains from each step in our proposed context-preserving negative caption construction.}
\label{tab:lang_model}
\resizebox{\columnwidth}{!}{%
\begin{tabular}{llcccc}
\toprule\noalign{\smallskip}
\bf Negative Captions & \bf Language Model & \bf R@1 & \bf R@5 & \bf R@10 & \bf Accuracy \\
\noalign{\smallskip}
\midrule
\noalign{\smallskip} 
None    & \texttt{BERT (Random)} & 25.66 & 59.57 & 75.16 & 57.37 \\
None    & \texttt{BERT (Pretrained)}  & 35.74 & 72.91 & 82.07 & 66.89 \\
\midrule
Random  & \texttt{BERT (Pretrained)} & 36.32 & 72.42 & 81.81 & 66.92 \\
Contextually plausible & \texttt{BERT (Pretrained)} & \sbest{48.05} & \sbest{76.78} & \sbest{82.97} & \sbest{74.91} \\
Excluding near-synonyms \& hypernyms  & \texttt{BERT (Pretrained)} & \best{51.67} & \best{77.69} & \best{83.25} & \best{76.74} \\
\bottomrule
\end{tabular}}
\end{center}
\end{table}
\setlength{\tabcolsep}{1.4pt}

%% file: figs/infonce_acc_plot.tex
\begin{figure}[t]
    \centering
    \includegraphics[height=6cm]{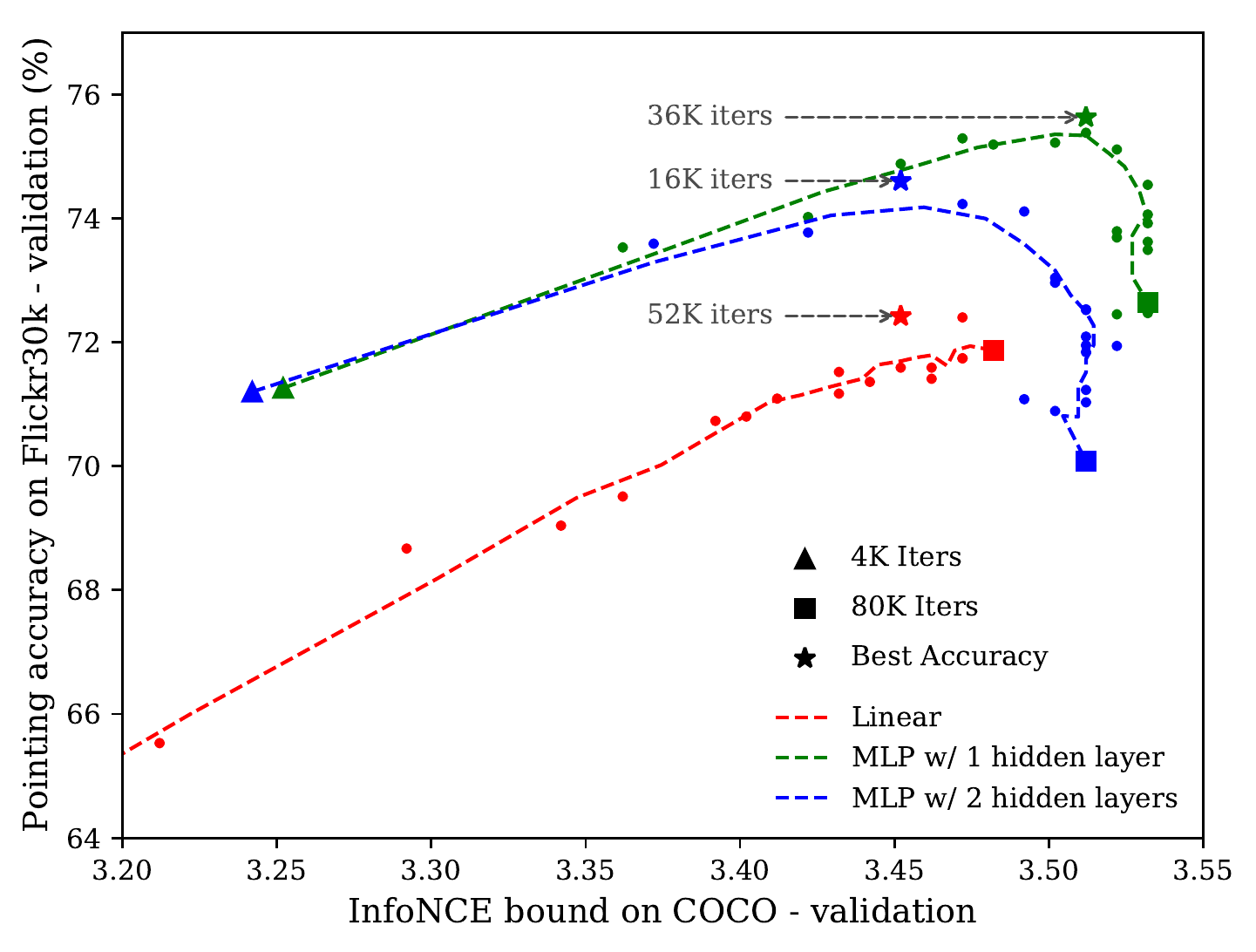}
    \vskip -6pt
    \caption{Relation between \texttt{InfoNCE} lower bound and phrase grounding performance with training iterations for 3 different choices of \texttt{key-value} modules in the compatibility function $\phi_{\theta}$. Each epoch is $\sim8$K iterations. The scattered points visualize the measured quantities during training. The dashed lines are created by applying moving average to highlight the trend.}
    \label{fig:infonce_acc_plot}
\end{figure}

%% file: figs/qual_results.tex
\begin{figure}[h]
    \centering
    \includegraphics[width=\linewidth, trim={0cm, 0, 0, 0cm}, clip=True]{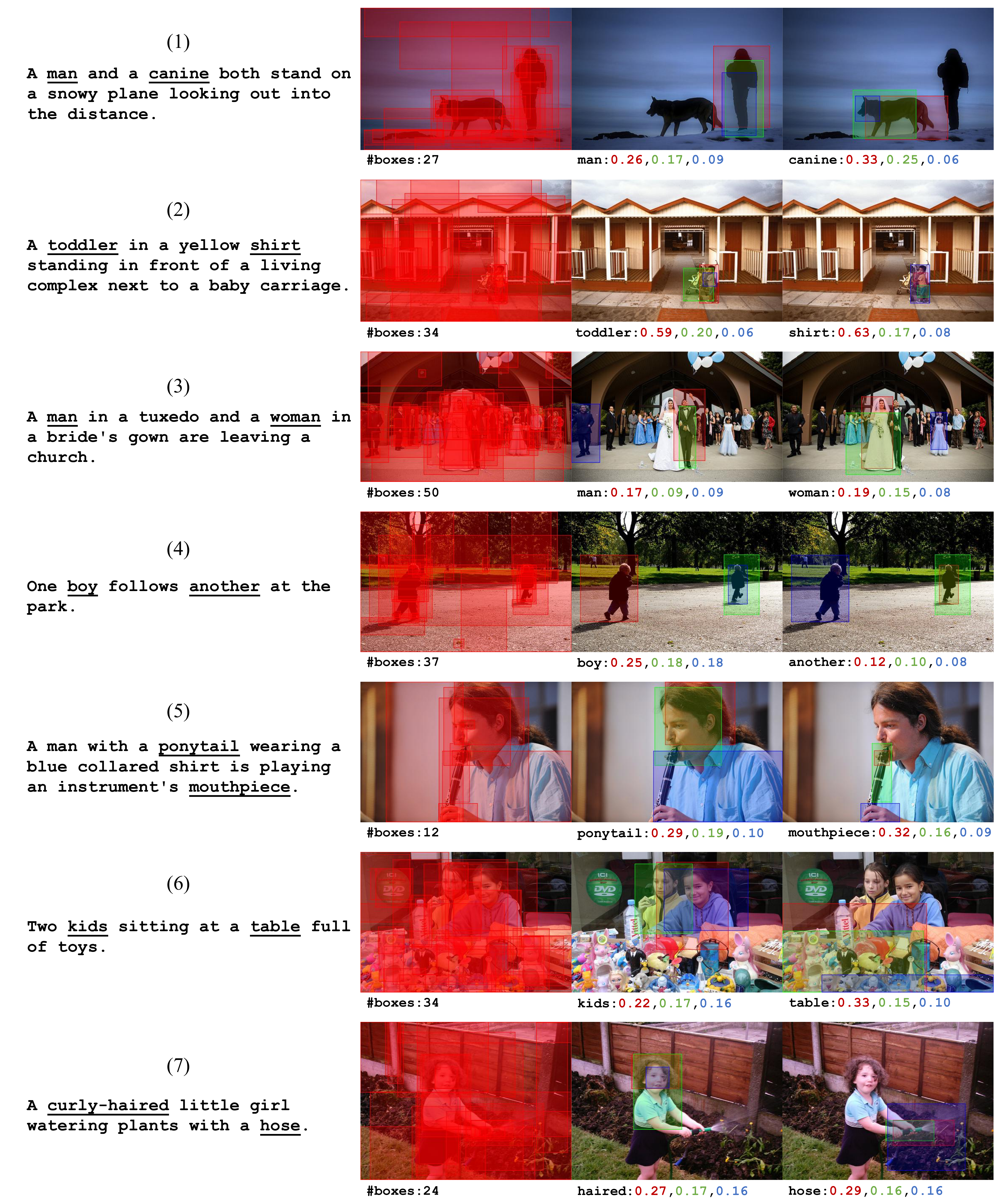}
    \caption{\textbf{Visualization of attention.} We show all detected regions and top-3 attended regions with attention scores for two words highlighted in each caption. More qualitative results can be found on our project page \url{http://tanmaygupta.info/info-ground/}}
    \label{fig:qual_results}
\end{figure}

%% file: appendix.tex
\section{Appendix}

\subsection{Limitations and Future Works}~\label{appendix:limitations}
The empirical examination of our framework reveals the following limitations:

\myparagraph{Pretrained representations.} Like prior arts, our approach relies on pretrained object detector and a language model to represent regions and caption-words. Ideally, we would expect to learn from scratch or improve existing region and word representations directly from image-caption data.

\myparagraph{Need for fully-labeled validation set.} In Fig.~\ref{fig:infonce_acc_plot}, we observe that an early stopping based on the validation performance is required to choose the best model for phrase grounding. While this is common practice for weakly supervised learning~\cite{choe2020evalWS} and the Flickr30K Entities validation set we use is $80\times$ smaller than the COCO training set, this translates to using full supervision for a small set of images. 

\myparagraph{Bounds on MI.} While $\log (K) - \mathcal{L}_{\texttt{img}}$ in Eq.~\ref{eq:infonce_grounding_loss} is a valid lower bound on MI, our $\log (K) - \mathcal{L}_{\texttt{lang}}$ in Eq.~\ref{eq:neg_loss} is no longer a lower bound on MI as it oversamples negative words related to a caption. A valid bound would involve random sampling of captions from the training data however our context-preserving negative captions lead to much better performance.

\subsection{Advantages of Context-Preserving Negative Sampling}

Commonly used strategies for negative sampling for contrastive learning include randomly sampling captions from the training data as negatives or mining hard-negatives from a randomly sampled mini-batch. In our experiments (Tab.~\ref{tab:lang_model}), random sampling showed no significant gains over a model trained without negative captions. This is because the sampled negatives often have an entirely different context as compared to the image and the positive caption which makes it too easy for the model to produce a low compatibility score for these negatives.

In contrast, contrast-preserving negative sampling shows significant gains over random sampling ($76.74\%$ \textit{vs.} $66.89\%$ pointing accuracy). This is because we construct harder negative captions which yield a more informative training signal than random sampling. We construct negatives by substituting only a single word in the caption while preserving the context from the positive caption. The substitutions are further chosen to be plausible given the context while discarding likely synonyms and hypernyms. Unlike random sampling approaches whose success depends on the occurrence of informative negative captions in the training data and the likelihood of sampling such negatives for a positive caption in the same minibatch, our approach can construct effective negatives for any positive caption.  

\subsection{Relation between our query-key-value attention and self-attention in Transformers}

Our query-key-value attention mechanism is related to the attention mechanism used in transformer-based~\cite{vaswani2017transformer} architectures like BERT~\cite{devlin2018bert}. Transformers use the mechanism for self-attention where queries, keys, and values are computed for each word in the input sentence and the attention scores are used for contextualization. In contrast, we use the attention mechanism for word-region alignment. Specifically, we compute queries for each contextualized word, keys for each region, and values for regions as well as words (using separate value networks for regions and words).

\subsection{Comparison to Align2Ground}
While we use the same visual features as the previous SOTA, Align2Ground~\cite{datta2019align2ground}, the two approaches use different textual features. While Align2Ground uses a bi-GRU, we chose BERT, a transformer-based language model which became more prevalent (as opposed to RNN-based) in the vision-language community. To estimate the gain due to pretrained language representations, Tab.~\ref{tab:lang_model} compares the grounding performance of randomly initialized BERT ($57.37\%$) to that of pretrained BERT ($66.89\%$). Negative sampling brings further gains ($76.74\%$).